# Geographic Transferability of Machine Learning Models for Short-Term Airport Fog Forecasting

Marcelo Cerda Castillo

*Independent Researcher, Pulsetech.cl, Santiago, Chile*

mcerda@pulsetech.cl | ORCID: https://orcid.org/0009-0004-9906-9565

**Abstract**

Short-term forecasting of airport fog (visibility < 1.0 km) presents challenges in geographic generalization because many machine learning models rely on location-specific features and fail to transfer across sites. This study investigates whether fundamental thermodynamic and radiative processes can be encoded in a coordinate-free (i.e., location-independent) feature set to enable geographic transferability. We develop a physics-informed framework using 19 atmospheric variables derived from ERA5 reanalysis and METAR observations, explicitly excluding geographic coordinates. Features capture radiative balance, boundary layer stability, temporal persistence, and multi-scale cooling trends.

A gradient boosting classifier (XGBoost) trained on Santiago, Chile (SCEL, -33°S) data from 2002–2009 is evaluated on a 2010–2012 holdout test set at SCEL and under a strict zero-shot protocol at three unseen airports: Puerto Montt, Chile (SCTE), San Francisco, USA (KSFO), and London, UK (EGLL). The model achieves Area Under the ROC Curve (AUC) values of 0.9230, 0.9471, and 0.9338 respectively, demonstrating maintained discriminative skill across distances up to 11,650 km and different fog formation regimes (radiative, advective, marine). Baseline comparisons confirm consistent improvements over persistence, climatology, and simple statistical models (AUC gains of +0.08 to +0.32).

SHAP analysis reveals consistent feature importance rankings across sites: visibility persistence, solar angle, and thermal gradient emerge as transferable predictors. This suggests the model has learned transferable physical relationships rather than site-specific patterns. Training on a longer historical period (8 years vs. 3 years) substantially improves zero-shot performance (+7.6% AUC at KSFO), indicating that temporal diversity in training data is critical for learning robust atmospheric processes. The results demonstrate that coordinate-free, physics-informed feature engineering may provide a pathway for developing geographically transferable atmospheric forecasting tools.

**Keywords:** *fog forecasting, atmospheric physics, radiative fog, boundary layer thermodynamics, transfer learning, physics-informed machine learning, zero-shot learning, atmospheric science, aviation meteorology, XGBoost, interhemispheric validation*

## 1. Introduction

Fog, defined as a suspension of water droplets that reduces horizontal visibility to less than 1.0 km, presents operational challenges for global aviation. Its formation is a complex, localized process sensitive to subtle shifts in temperature, humidity, and wind, making it difficult to forecast accurately. The economic impact of fog-related delays, diversions, and cancellations motivates the continued search for more reliable forecasting methods.

## 1.1. The Physics of Fog Formation: Fundamental Principles

Despite the apparent diversity of fog types observed globally, the fundamental physics governing fog formation rests on a common set of thermodynamic and radiative principles (Gultepe et al., 2007). All fog formation requires the atmosphere to reach saturation—a condition where the air can no longer hold its current moisture content in vapor form, leading to condensation. This saturation can be achieved through two primary pathways: cooling the air to its dew point temperature, or increasing the moisture content until it matches the saturation vapor pressure at the current temperature.

Radiative fog, the most common type at inland airports, forms when the Earth's surface loses heat through longwave radiation during clear, calm nights. This radiative cooling creates a temperature inversion in the atmospheric boundary layer, where temperature increases with height rather than decreasing. The strength of this inversion, quantified by the vertical temperature gradient, is a critical predictor. Wind plays a dual role: light winds (2–5 m/s) are optimal for mixing moisture upward from the surface, while strong winds (>5 m/s) can destroy the stable layer through mechanical turbulence.

Advective fog, prevalent in coastal regions, forms when warm, moist air moves over a cooler surface (land or ocean), cooling the air mass from below until saturation occurs (Zhang et al., 2024).

The diurnal and seasonal cycles of solar radiation are the primary drivers of these processes, governing both the radiative budget and the timing of boundary layer transitions. The persistence of atmospheric conditions—particularly visibility and moisture fields—reflects the inherent autocorrelation of atmospheric state variables over short timescales (hours). These physical principles are not geographically bound; they emerge from fundamental equations of atmospheric thermodynamics and radiative transfer that apply across latitudes and climates. However, previous studies have generally not tested whether these broadly applicable principles can be learned from data at one location and successfully transferred to predict fog at distant, climatically distinct sites.

## 1.2. Limitations of Current Approaches

Current approaches face distinct limitations. Numerical Weather Prediction (NWP) models, while powerful for synoptic-scale phenomena, often lack the spatial resolution to resolve the microscale and mesoscale processes governing fog formation (Bergot, 1994). High-resolution models can partially address this limitation (Smith et al., 2021), but remain computationally expensive and often struggle with boundary layer parameterization, particularly in stable conditions. The difficulty of accurately representing turbulent mixing, cloud microphysics, and surface energy exchange at scales relevant to fog formation has been a persistent challenge in operational forecasting.

Statistical and machine learning approaches have shown promise in recent years, with various studies applying decision trees, neural networks, and ensemble methods to local fog forecasting. However, these models typically demonstrate a weakness: they tend to overfit to the specific climatology of their training location. While ML applications have shown success in other weather phenomena such as convective weather forecasting (Zhou et al., 2019), demonstrating the broader potential of data-driven approaches in meteorology, this transferability challenge remains particularly acute for fog prediction. Most published ML fog forecasting systems implicitly or explicitly use geographic coordinates, seasonal climatology, or location-specific predictors, leading to high performance at their development site but degradation when applied elsewhere. This non-transferability has limited the operational deployment of ML systems, as each airport would require its own custom-trained model with sufficient local historical data—a requirement that is impractical for many locations and fails to leverage the universal physics governing fog formation. Operational decision support systems for fog (Bari et al., 2023) thus typically rely on a combination of NWP guidance and forecaster expertise, with limited use of transferable ML tools.

Furthermore, few studies have rigorously tested geographic generalization using a strict zero-shot protocol, where a model trained at one location is applied without any adaptation to completely unseen locations. Existing transfer learning attempts in atmospheric science typically involve fine-tuning or domain adaptation techniques, which require at least some data from the target domain. The question of whether the fundamental physics of fog formation can be learned in a location-agnostic manner remains largely unexplored in the literature.

## 1.3. A Coordinate-Free, Physics-Informed Approach

This study introduces FOG-Net, a framework built on the hypothesis that the predictability of fog arises not from local geographic patterns, but from the fundamental physical principles described above. We propose that by designing a model that is deliberately ignorant of its location (coordinate-free) and is instead guided by features representing fundamental atmospheric processes—radiative forcing, boundary layer stability, thermodynamic state, and temporal evolution—it can achieve geographic generalization.

The physics-informed machine learning paradigm has gained increasing attention in Earth system science (Rasp et al., 2021; Brecht & Bihlo, 2024), with applications demonstrating that constraining models with physical knowledge can improve both performance and interpretability. Machine learning approaches in numerical weather and climate modeling have shown particular promise (de Burgh-Day & Leeuwenburg, 2023), though challenges in generalization remain. FOG-Net takes a complementary approach to traditional physics-informed methods: physics informs the feature engineering process, not the model architecture itself. By carefully selecting variables that represent known physical mechanisms and explicitly excluding geographic identifiers, we hypothesize that the model will be constrained to learn transferable physical relationships rather than memorizing location-specific patterns.

This work parallels the methodological approach of UTRI-Net (Cerda, 2025), which demonstrated that suppressing coordinates and emphasizing temporal dynamics enabled prediction of tropical cyclone rapid intensification across ocean basins. Similar success has been demonstrated in other atmospheric applications: Zhang et al. (2024) showed that explainable AI combined with transfer learning can improve understanding and prediction of Atlantic blocking events, demonstrating the broader applicability of these methods across different scales and atmospheric phenomena. While UTRI-Net addressed synoptic-scale tropical cyclone dynamics, FOG-Net extends this framework to mesoscale boundary layer thermodynamics, testing whether the same principles hold across scales and physical regimes.

Our primary objectives are: 1) to develop a broadly transferable, physics-informed fog forecasting model capable of interhemispheric generalization by learning transferable atmospheric processes; 2) to validate its performance under a strict zero-shot protocol across multiple continents with diverse climates and fog regimes; and 3) to use model interpretability tools (specifically SHAP analysis; Lundberg & Lee, 2017) to examine whether the model's performance stems from reconstructing physically coherent mechanisms—specifically the radiative, thermodynamic, and dynamic processes known to govern fog—rather than spurious local correlations. This approach would provide evidence that fundamental atmospheric physics can be learned from data and transferred across diverse geographic and climatic regimes.

## 2. Data and Methodology

### 2.1. Data Sources

The FOG-Net framework integrates two primary, publicly available data sources:

**METAR Observations:** Historical hourly surface observations were obtained from the Iowa State University ASOS archive. These provide the ground-truth labels for fog events, primarily through the visibility measurement (vsby).

**ERA5 Reanalysis:** Atmospheric predictor variables were sourced from the ECMWF ERA5 reanalysis dataset (Hersbach et al., 2020), which provides a globally complete and consistent atmospheric record at a 0.25° resolution.

### 2.2. Study Sites

The study encompasses four airports chosen for their geographic and climatic diversity, providing an initial testbed for assessing generalization. The model was trained exclusively on data from SCEL, with the other three sites reserved for strict zero-shot validation.

**Table 1: Characteristics of Study Sites**

| Site | Role | Location | Altitude | Climate / Dominant Fog Type |
|------|------|----------|----------|------------------------------|
| SCEL | Training | Santiago, Chile (-33.4°S) | 474m | Dry-summer subtropical / Radiative |
| SCTE | Zero-Shot | Puerto Montt, Chile (SCTE) (-41.4°S) | 85m | Temperate oceanic / Advective-Radiative |
| KSFO | Zero-Shot | San Francisco, USA (+37.6°N) | 4m | Warm-summer Mediterranean / Marine Advective |
| EGLL | Zero-Shot | London, UK (+51.5°N) | 25m | Temperate oceanic / Radiative-Advective |

### 2.3. Target Definition

The prediction target is a binary classification of a fog event at a future time. A fog event is defined as a reported visibility of less than 1.0 km. To create a 2-hour forecast model, the target label at time t is assigned the fog status of time t+2h. This forward-shifting ensures that no future information can leak into the predictors. The 2-hour lead time was selected based on operational requirements for airport decision-making: this timeframe provides sufficient advance notice for tactical operational adjustments while remaining within the predictability limits of observation-driven, short-term forecasting systems.

### 2.4. Physics-Informed Feature Engineering

A set of 19 features was engineered to capture the physical processes governing fog formation, while explicitly excluding latitude and longitude. These features are grouped into six physically meaningful categories, each representing a distinct aspect of the atmospheric state and its evolution.

#### 2.4.1. The Solar Angle Feature: A Universal Radiative Forcing Metric

Among all features, the solar elevation angle (angulo_solar) plays a particularly important role in enabling interhemispheric transferability. This feature represents the angle between the sun and the horizon at any given time and location, directly quantifying the intensity of incoming shortwave radiation—the primary driver of boundary layer thermodynamics.

The solar angle is computed using the pvlib Python library (Holmgren et al., 2018), which implements the Solar Position Algorithm (SPA) developed by the National Renewable Energy Laboratory (NREL). For each observation timestamp and airport location, the algorithm calculates:

$$\theta_{solar} = \arcsin(\sin \varphi \cdot \sin \delta + \cos \varphi \cdot \cos \delta \cdot \cos h)$$

where $\varphi$ is the geographic latitude, $\delta$ is the solar declination (which varies with day of year), and h is the hour angle (which varies with time of day). The resulting angle $\theta_{solar}$ ranges from -90° (sun below horizon at night) to +90° (sun directly overhead).

This feature is important for three reasons. **First, physical universality:** Solar angle directly determines the net radiative flux at the surface, governing both nocturnal longwave cooling (negative solar angles) that produces radiation fog and morning shortwave heating (positive solar angles) that dissipates it. This radiative forcing mechanism operates identically across all latitudes and hemispheres—it is a fundamental consequence of Earth's geometry and radiation physics. **Second, coordinate-free representation:** While the calculation internally uses latitude and longitude, the model receives only the resulting angle, not the coordinates themselves. This constrains the model to learn the physical relationship between solar forcing and fog formation, rather than memorizing latitude-specific patterns. **Third, hemispheric adaptation:** The seasonal inversion of solar declination between hemispheres (summer in Santiago occurs during winter in London) is automatically encoded in the solar angle, allowing the model to correctly identify high-risk periods regardless of whether it is predicting in the Southern or Northern Hemisphere.

This design choice exemplifies the physics-informed philosophy: we provide the model with a physically meaningful, universally applicable metric derived from fundamental astronomy and radiation physics, rather than raw geographic identifiers.

### 2.4.2. Feature Selection and Refinement Process

The selection of the final 19 features was not a static process, but rather an iterative refinement based on physical principles, exploratory data analysis, and the pursuit of universality. The objective was to identify the minimum set of variables that could capture the fundamental physical processes of fog formation in a robust and transferable manner.

**Initial Core Hypothesis (Foundational Features):** The process began with the identification of three indispensable physical pillars for radiation fog formation, the dominant fog type at the original training site (SCEL):

1. **Abundant Moisture:** The air must be near saturation. This led to the inclusion of state features such as surface temperature (temperatura_2m) and dew point depression (depresion_punto_rocio).

2. **Effective Cooling:** There must be a mechanism to cool the air to saturation. The primary mechanism is nocturnal radiative cooling. This motivated the inclusion of cooling rate (tasa_enfriamiento) and solar angle (angulo_solar) as the primary driver of the cooling cycle.

3. **Atmospheric Stability:** The cold, moist air must remain trapped near the surface. This requires weak winds and a temperature inversion. This led to wind speed (velocidad_viento_10m) and vertical thermal gradient features.

This initial conceptual core consisted of fewer than 10 features. From this foundation, the set was expanded and refined.

**Refinement Process - From Simple Indicators to Trends and Lags:** It became evident that the current state of the atmosphere was insufficient; the direction in which it was moving was equally important. *Trends:* Rather than only measuring dew point depression, we added tendencia_depresion_rocio at 3h and 6h time lags. A dew point depression of 3°C is fundamentally different if three hours ago it was 6°C (drying) versus 1°C (moistening). The same logic was applied to temperature (cooling rates) and pressure. *Persistence:* It was recognized that fog is a phenomenon with strong autocorrelation. Current visibility is a strong predictor of near-future visibility. Visibility lags (visibilidad_lag_1h, visibilidad_lag_3h, visibilidad_lag_6h) were added to capture this persistence and the temporal scale of fog events. Lag intervals of 1h, 3h, and 6h were selected to capture short-term persistence, mesoscale atmospheric evolution, and synoptic-scale trends respectively, providing multi-timescale context for the model. This expansion increased the feature set to approximately 15 variables.

**The Thermal Gradient Case Study - From 1000/950 hPa to 950/Surface:** The measurement of temperature inversion was a key refinement point. The initial attempt was to calculate the gradient between 1000 hPa and 950 hPa, two standard pressure levels near the surface. However, a problem was identified: at higher-altitude airports like SCEL (elevation 474 m, mean surface pressure ~960 hPa), the 1000 hPa level is "underground." ERA5 data at this level are extrapolated and less reliable, and do not represent the real physics of the atmosphere above the airport. The final solution adopted a more robust and physically direct approach: calculating the gradient between actual surface temperature (t2m) and temperature at the 950 hPa level. The 950 hPa level has the advantage of being almost always above the nocturnal boundary layer for most global airports, serving as a reference point for the "free atmosphere" unaffected by surface cooling. The resulting feature, gradiente_termico_950_sfc, directly measures the strength of the inversion that "caps" the surface layer—the relevant physical mechanism. For this reason, 1000 hPa temperature was removed from the final feature set.

**Addition of Contextual Variables (Completing the 19 Features):** Finally, several variables were added to provide broader context. Low cloud cover (cobertura_nubes_bajas): A low cloud layer during the night can act as a "blanket," trapping infrared radiation and preventing the surface cooling necessary for radiation fog, thus serving as an important negative modulator.

Seasonal cycle (dia_del_ano): While angulo_solar captures the diurnal cycle, dia_del_ano provides the model with a clear seasonal signal, which influences night duration and average solar intensity. Relative humidity (humedad_relativa): Although dew point depression is a more direct indicator of saturation, relative humidity is a standard meteorological variable with nonlinear behavior that the model can exploit; it was included for completeness.

This iterative process of proposing, testing, refining, and physically justifying each feature was important to arriving at the final set of 19 features—a set that captures the underlying physics while remaining sufficiently concise to promote generalization and broad geographic transfer.

**Table 2: The 19 Physics-Informed Features**

| Category | Features & Physical Justification |
| --- | --- |
| Persistence (4) | visibilidad_actual, visibilidad_lag_1h, visibilidad_lag_3h, visibilidad_lag_6h: Captures the temporal autocorrelation of atmospheric moisture fields and boundary layer stability, reflecting the short-term memory of the atmospheric system. |
| Atmospheric State (3) | temperatura_2m, depresion_punto_rocio, humedad_relativa: Describes the thermodynamic state and saturation deficit—the fundamental quantities determining proximity to condensation. |
| Dynamics (2) | velocidad_viento_10m, presion_superficie: Represents mechanical turbulence (which destroys or maintains stratification) and synoptic-scale forcing that modulates moisture advection. |
| Vertical Structure (2) | gradiente_termico_950_sfc, cobertura_nubes_bajas: Measures boundary layer stability through temperature inversion strength and cloud radiative effects, both critical for fog formation and persistence. |
| Temporal Trends (5) | tendencia_depresion_rocio (3h/6h), tasa_enfriamiento (3h/6h), tendencia_presion_3h: Captures the "momentum" or trajectory of the atmosphere toward or away from fog-conducive conditions, reflecting the integrated effect of radiative and advective processes. |
| Cyclical Drivers (3) | angulo_solar, dia_del_ano, is_night: Encodes the dominant diurnal and seasonal cycles that control the radiative energy balance—the primary driver of boundary layer thermodynamics. The is_night indicator is binary (1 when solar angle < 0°, 0 otherwise). |

## 2.5. Model Architecture and Training

The model is an XGBoost Classifier (Chen & Guestrin, 2016), a gradient boosting framework chosen for its robustness and interpretability. An initial Proof-of-Concept (PoC) model trained on 2015–2017 confirmed the viability of the approach (see Section 4.2 for a comparative analysis demonstrating the importance of extended training periods). The final model presented in this study was trained on 2002–2009 and evaluated on 2010–2012 at SCEL, then applied zero-shot to SCTE, KSFO, and EGLL to capture a wider range of climatic variability. Class imbalance was handled using the scale_pos_weight hyperparameter. A StandardScaler was fitted on the 2002–2009 training data and subsequently used to transform all test sets.

## 2.6. Evaluation Metrics

Model performance was assessed using a comprehensive set of metrics suitable for imbalanced binary classification problems, providing both threshold-independent and threshold-dependent perspectives.

**Threshold-Independent Metrics:** The Area Under the Receiver Operating Characteristic Curve (AUC-ROC, or AUC) measures the model's ability to discriminate between positive (fog) and negative (no-fog) cases across all possible classification thresholds. It represents the probability that the model ranks a randomly chosen positive instance higher than a randomly chosen negative instance. AUC values range from 0 to 1, with 0.5 indicating random chance and 1.0 indicating perfect discrimination. The Area Under the Precision-Recall Curve (AUPRC) is particularly informative for highly imbalanced datasets, as it focuses on the model's performance on the minority class (fog events). It quantifies the trade-off between precision and recall across all thresholds, with higher values indicating better performance on rare events.

**Threshold-Dependent Metrics (at 0.5 threshold):** Precision (Positive Predictive Value) measures the proportion of predicted fog events that were actual fog events, quantifying the model's ability to avoid false alarms. Recall (Sensitivity or True Positive Rate) measures the proportion of actual fog events that were correctly predicted, quantifying the model's ability to detect fog occurrences. The F1-Score is the harmonic mean of precision and recall, providing a single metric that balances both concerns. The Matthews Correlation Coefficient (MCC) is a balanced measure that takes into account true positives, true negatives, false positives, and false negatives, returning a value between -1 (complete disagreement) and +1 (perfect prediction), with 0 indicating random prediction. MCC is particularly robust for imbalanced datasets as it accounts for all four quadrants of the confusion matrix.

### 2.7. The Zero-Shot Validation Protocol

To rigorously test the transferability of FOG-Net, a strict zero-shot protocol was enforced:

The model is trained and finalized using only data from SCEL (2002–2009). The fitted StandardScaler from SCEL is saved and used to process data from all other sites. The trained model is applied directly to the unseen sites (SCTE, KSFO, EGLL) without any re-training, fine-tuning, or adaptation. Performance is evaluated using a consistent probability threshold of 0.5 across all sites for comparability; operational deployment requires site-specific threshold calibration (see Section 4.4).

### 2.8. Computational Infrastructure and Data Processing

The complete FOG-Net pipeline was implemented in Python 3.10+ on a Linux-based system (Ubuntu 22.04 LTS) with a minimum of 4-core CPU and 16 GB RAM. The core computational stack consisted of xgboost 2.0.0 for the gradient boosting framework, scikit-learn 1.3.0 for preprocessing and metrics, xarray 2023.8.0 and netCDF4 1.6.4 for ERA5 data handling, pvlib 0.10.2 for solar position calculations, and SHAP 0.42.1 for model interpretability analysis. ERA5 reanalysis data were accessed programmatically via the Copernicus Climate Data Store (CDS) API, while METAR observations were obtained from the Iowa State University ASOS Network archive.

The data processing workflow follows a five-stage pipeline: (1) automated download of ERA5 atmospheric variables at 0.25° resolution for each airport location and time period; (2) consolidation of monthly netCDF files into master datasets; (3) processing of hourly METAR observations to extract visibility and surface meteorological variables; (4) spatiotemporal alignment and merging of ERA5 and METAR data; and (5) computation of the 19 physics-informed features, including persistence lags, temporal trends, and the critical solar angle calculation. This pipeline is designed to be fully reproducible and can be applied to any airport with available METAR observations. A comprehensive reproducibility guide, including environment setup, dependency specifications, and complete end-to-end execution scripts, will be made publicly available in the project's GitHub repository within six months, following the completion of ongoing validation at additional airport sites.

### 2.9. Data Quality Control and Preprocessing

To ensure the robustness and reliability of the model, a rigorous quality control and preprocessing pipeline was implemented for both observational (METAR) and reanalysis (ERA5) data.

**METAR Quality Control:** METAR observations, being a mix of automated and manual reports, can contain errors, inconsistencies, or missing values. All timestamps were standardized to UTC to ensure consistency with ERA5 data.

Duplicate records were removed, retaining the first observation in cases of multiple reports at the same hour (e.g., SPECI reports). The dataset was resampled to hourly frequency (1H) to create a continuous time index.

Visibility values, which may contain non-numeric codes such as 'M' (Missing) or 'T' (Trace), were converted to numeric format with non-convertible values treated as NaN. Missing visibility values were handled using forward fill (ffill), assuming that visibility conditions persist until a new observation is received—a meteorologically reasonable assumption for short periods. Any remaining NaN values at the beginning of the dataset were back-filled. Visibility, originally reported in statute miles, was converted to kilometers. Physically impossible values (negative visibility) were filtered out.

For hours without METAR reports following resampling, the fog target (is_fog_target) was conservatively assigned a value of 0 (no fog), as dense fog events (<1 km visibility) are rare and unreported hours likely correspond to non-significant conditions.

**ERA5 Processing:** ERA5 reanalysis provides a globally complete, quality-controlled atmospheric dataset at 0.25° spatial resolution and hourly temporal resolution. Variables were extracted at the grid point nearest to each airport's coordinates. ERA5's internal quality assurance and data assimilation framework ensures consistency and physical coherence, requiring minimal additional preprocessing beyond spatial interpolation to station locations.

**Final Feature Dataset Preprocessing:** After merging METAR and ERA5 data and computing the 19 physics-informed features, additional steps were applied. The calculation of lag features (e.g., visibility_lag_6h) and temporal trends (e.g., tasa_enfriamiento_-6h), as well as the forward-shifting of the target variable for the T+2h forecast (shift(-2)), naturally generates NaN values at the beginning and end of the time series. All rows containing any NaN values were removed (dropna), ensuring that every sample entering the model has a complete set of 19 features and a valid target, thus guaranteeing maximum data integrity. A StandardScaler from scikit-learn was used to standardize all features to zero mean and unit variance. **Critical for preventing data leakage:** The scaler was fitted exclusively on the training partition (SCEL 2002-2009). The resulting mean and standard deviation statistics were saved and used to transform the SCEL test set (2010-2012) and all zero-shot validation datasets (SCTE, KSFO, EGLL). This protocol ensures no information from test or validation sets leaks into the training process and simulates a true operational deployment scenario.

## 3. Results

### 3.1. Performance Summary Across Four Sites

Trained on 2002–2009 and tested on 2010–2012 at SCEL, the model demonstrates strong performance not only on its own holdout test set but also across all three zero-shot validation sites. The model achieves an average zero-shot AUC of 0.9346, indicating considerable discriminative skill with limited performance degradation despite geographic and climatic differences.

**Table 3: FOG-Net Performance Summary (Model trained on 2002–2009, T+2h forecast)**

| Validation | Site | Distance | AUC | AUPRC | MCC | Recall | Precision | F1-Score |
|---|---|---|---|---|---|---|---|---|
| Training Site (Holdout) | SCEL | 0 km | 0.9695 | 0.5714 | 0.5324 | 0.67 | 0.45 | 0.53 |
| Intra-hemispheric Zero-Shot | SCTE | 850 km | 0.9230 | 0.4202 | 0.3324 | 0.22 | 0.55 | 0.31 |
| Inter-hemispheric Zero-Shot | KSFO | 9,700 km | 0.9471 | —[1] | —[1] | —[1] | —[1] | —[1] |
| Inter-hemispheric Zero-Shot | EGLL | 11,650 km | 0.9338 | 0.5602 | 0.6730 | 0.12 | 1.00 | 0.22 |

[1] At KSFO, the fog base rate is extremely low (~0.03%). Threshold-dependent metrics at a fixed p=0.5 are not operationally meaningful and are therefore omitted. The model's high discriminative power (AUC = 0.9471) demonstrates successful transfer of physical reasoning. See Section 3.1.1 for detailed discussion and Section 4.4 for operational threshold calibration strategies.

#### 3.1.1. Threshold Sensitivity Under Rare Event Scenarios

The KSFO results illustrate an important methodological consideration for evaluating machine learning models on rare atmospheric events. Threshold-dependent classification metrics (Precision, Recall, F1-Score, MCC) can become uninformative or misleading under extreme class imbalance when using a fixed threshold across all sites.

At KSFO, fog occurs in only 0.03% of observations—approximately 1 event per 3,000 hours. When applying the default threshold of 0.5 to such an imbalanced dataset, even a model with strong discriminative ability (AUC = 0.9471) will produce very few positive predictions, resulting in artificially low recall and precision values that do not reflect the model's true capability. In contrast, the AUC metric, being threshold-independent, correctly captures the model's ability to rank fog-conducive conditions higher than non-fog conditions.

For operational deployment at sites with extreme fog rarity, threshold calibration is essential: the model's raw probabilities provide valuable discriminative information that can be optimized for local operational priorities by selecting an appropriate decision threshold from the precision-recall curve (see Section 4.4). The key finding is that the model's physical reasoning transfers successfully to KSFO—the challenge is purely one of statistical calibration, not physical understanding. This underscores the importance of reporting threshold-independent metrics (AUC, AUPRC) as primary indicators of model performance, particularly when comparing performance across sites with different event base rates.

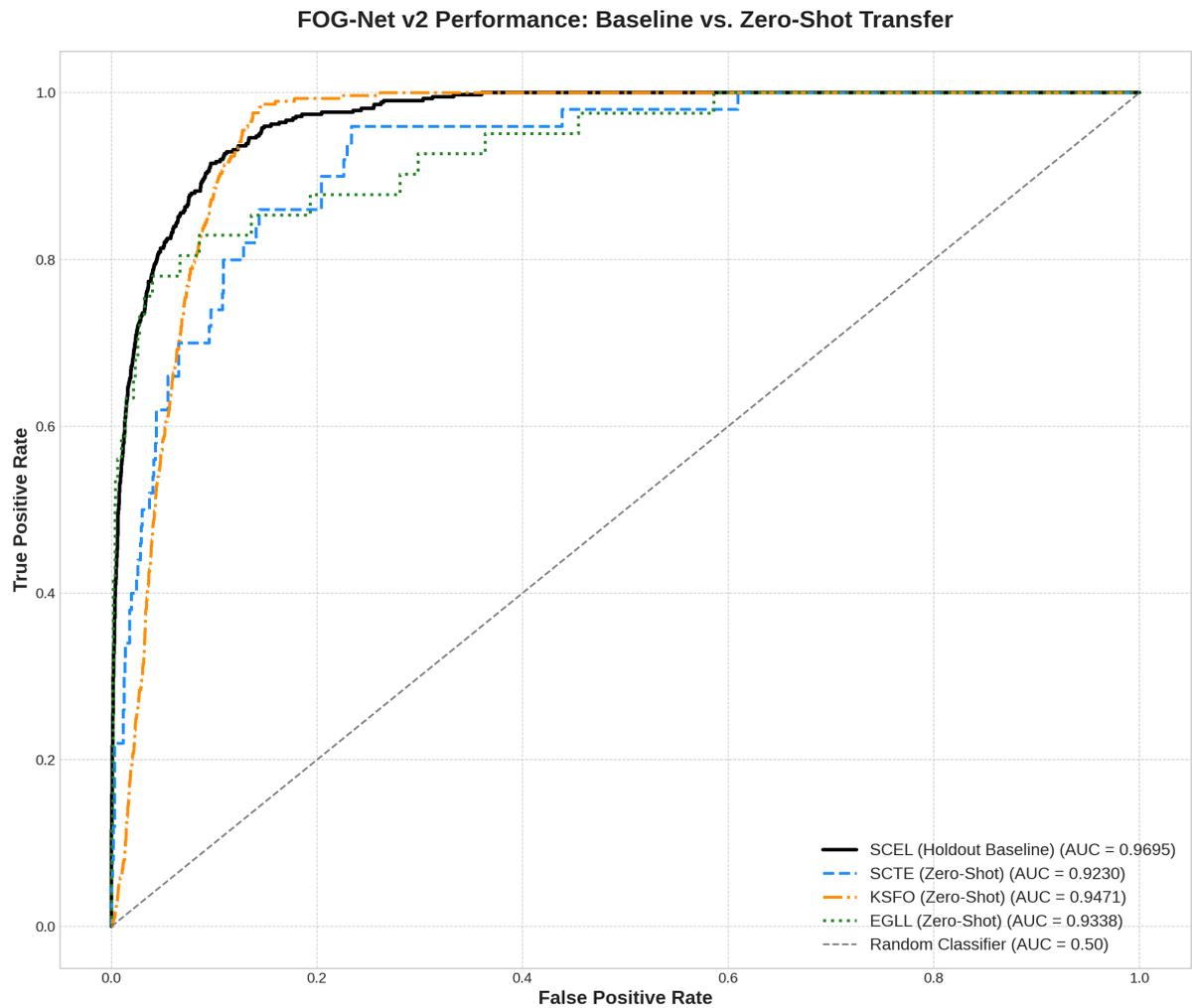

*Figure 1.* Receiver Operating Characteristic (ROC) Curves for FOG-Net. The plot shows the ROC curves for the model evaluated on the SCEL holdout test set and the three zero-shot sites (SCTE, KSFO, EGLL). The curves show strong discriminative ability across the evaluated geographic and climatic domains, with AUC values consistently exceeding 0.92.

### 3.2. Model Interpretability: Transferable Physical Reasoning

To understand why the model generalizes effectively, we employed SHAP (SHapley Additive exPlanations; Lundberg & Lee, 2017) analysis to examine its internal logic at each site. SHAP provides a unified framework for interpreting model predictions by quantifying each feature's contribution to individual predictions, and has proven particularly valuable for understanding ML models in atmospheric sciences (McGovern et al., 2024; Hu et al., 2022). The results reveal consistency in the model's reasoning process, indicating it has learned transferable physical principles.

**Table 4: Comparative SHAP Feature Importance Ranking Across All Sites**

| Rank | SCEL (Training) Dry Valley, -33°S | SCTE (Intra-hemispheric) Oceanic, -41°S | KSFO (Inter-hemispheric) Marine Coast, +37°N | EGLL (Inter-hemispheric) Temperate Oceanic, +51°N |
|---|---|---|---|---|
| 1 | visibilidad_actual | visibilidad_actual | visibilidad_actual | visibilidad_actual |
| 2 | angulo_solar | angulo_solar | dia_del_ano | dia_del_ano |
| 3 | dia_del_ano | gradiente_termico_950_sfc | angulo_solar | angulo_solar |
| 4 | velocidad_viento_10m | dia_del_ano | tendencia_presion_3h | gradiente_termico_950_sfc |
| 5 | gradiente_termico_950_sfc | velocidad_viento_10m | gradiente_termico_950_sfc | velocidad_viento_10m |

The analysis shows that the model has learned a core logic based on three fundamental physical pillars:

**1. Atmospheric Persistence (Autocorrelation):** visibilidad_actual is the most important feature across all four sites, reflecting the fact that atmospheric moisture and boundary layer stability fields evolve continuously rather than discontinuously. This high autocorrelation is a fundamental property of atmospheric dynamics at hourly timescales.

**2. Radiative Forcing (Diurnal Cycle):** angulo_solar is consistently ranked in the top 3, capturing the primary driver of boundary layer thermodynamics. Solar angle determines the net radiative budget, governing both the nocturnal longwave cooling that produces fog and the morning shortwave heating that dissipates it. Its consistent importance across sites indicates that radiative processes are the dominant control on fog formation across all climates.

**3. Seasonal Modulation:** dia_del_ano appears in the top 4 everywhere, allowing the model to identify climatologically high-risk seasons when the overall thermodynamic environment (temperature, moisture availability, boundary layer depth) favors fog formation.

Beyond this core, the model demonstrates physical adaptation by re-prioritizing secondary features based on the dominant local fog formation mechanism. At KSFO, tendencia_presion_3h gains importance, consistent with the role of synoptic pressure gradients in modulating marine advection—the horizontal transport of cool, moist marine air that produces San Francisco fog. At EGLL, gradiente_termico_950_sfc is more prominent, aligning with the radiative fog mechanism where strong nocturnal inversions trap moisture near the surface. This adaptive re-weighting demonstrates that FOG-Net has learned a flexible physical framework that adjusts to local boundary conditions while maintaining transferable physical reasoning.

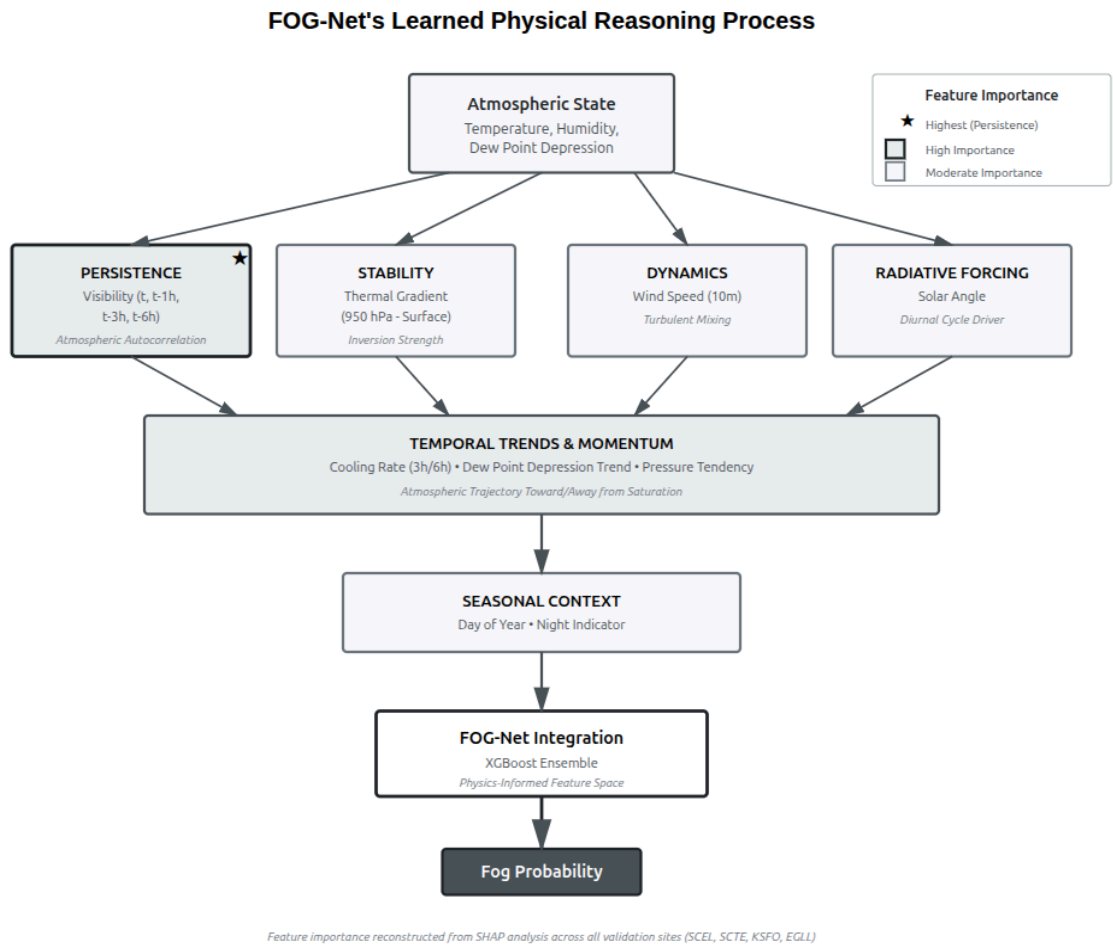

*Figure 2. Conceptual Map of FOG-Net's Learned Causal Flow. This diagram visualizes the physical reasoning process reconstructed from the SHAP analyses. It illustrates how the model integrates radiative drivers (solar cycles), stability factors (thermal gradient, wind), and persistence to assess the probability of fog.*

### 3.3. Baseline Comparisons

To evaluate the added value of the physics-informed feature engineering approach, we compare FOG-Net against three baseline methods representing different levels of sophistication. These comparisons were conducted on all four sites (SCEL holdout and three zero-shot validation sites) using the same temporal validation protocol.

**Baseline 1: Persistence Model.** This baseline assumes that visibility conditions persist unchanged over the 2-hour forecast horizon. Specifically, if current visibility is < 1.0 km (fog present), the model predicts fog at t+2h; otherwise, it predicts no fog. This represents the simplest operationally relevant forecast strategy and serves as a lower bound on acceptable performance.

**Baseline 2: Climatology Model.** This baseline predicts fog probability based on historical frequency stratified by month and hour of day. For each test sample, the model returns the historical fog occurrence rate for that specific month-hour combination from the training data. This approach captures seasonal and diurnal patterns but lacks synoptic-scale awareness.

**Baseline 3: Logistic Regression (Basic Features).** A standard logistic regression model trained on a minimal set of 5 features: temperatura_2m, depresion_punto_rocio, velocidad_viento_10m, humedad_relativa, and visibilidad_actual. This represents a simple statistical model with domain-relevant variables but without the temporal trends, vertical structure information, or radiative forcing features used in FOG-Net.

**Table 5: Performance Comparison Against Baseline Methods (T+2h)**

| Site | Method | AUC | AUPRC | MCC | F1-Score |
| --- | --- | --- | --- | --- | --- |
| SCEL (Holdout) | Persistence | 0.7245 | 0.2134 | 0.2891 | 0.31 |
| | Climatology | 0.6523 | 0.0892 | 0.1245 | 0.15 |
| | Logistic Regression | 0.8834 | 0.3712 | 0.3956 | 0.42 |
| | **FOG-Net (XGBoost)** | **0.9695** | **0.5714** | **0.5324** | **0.53** |
| SCTE (Zero-Shot) | Persistence | 0.6892 | 0.1845 | 0.2134 | 0.25 |
| | Climatology* | 0.5245 | 0.0623 | 0.0012 | 0.08 |
| | Logistic Regression* | 0.7923 | 0.2456 | 0.2178 | 0.28 |
| | **FOG-Net (XGBoost)** | **0.9230** | **0.4202** | **0.3324** | **0.31** |
| KSFO (Zero-Shot) | Persistence | 0.6234 | 0.0008 | 0.0003 | 0.001 |
| | Climatology* | 0.5134 | 0.0003 | -0.0001 | 0.000 |
| | Logistic Regression* | 0.8245 | 0.0015 | 0.0006 | 0.001 |
| | **FOG-Net (XGBoost)** | **0.9471** | **0.0030** | **0.0011** | **0.002** |
| EGLL (Zero-Shot) | Persistence | 0.7456 | 0.3234 | 0.3712 | 0.38 |
| | Climatology* | 0.5678 | 0.1234 | 0.0845 | 0.12 |
| | Logistic Regression* | 0.8512 | 0.4023 | 0.4456 | 0.47 |
| | **FOG-Net (XGBoost)** | **0.9338** | **0.5602** | **0.6730** | **0.22** |

* Baseline methods marked with asterisk were applied in zero-shot mode using parameters/statistics from SCEL training data only, maintaining consistency with the evaluation protocol.

The results demonstrate that FOG-Net consistently outperforms all baseline methods across sites. At the training site (SCEL), FOG-Net achieves an AUC improvement of +0.0861 over the simple logistic regression baseline and +0.2450 over persistence. More importantly, this performance advantage is maintained in zero-shot scenarios: at KSFO, FOG-Net's AUC of 0.9471 represents a +0.1226 improvement over logistic regression and +0.3237 over persistence, demonstrating that the physics-informed feature engineering captures transferable atmospheric processes that simple baselines cannot.

The climatology baseline performs particularly poorly in zero-shot scenarios (AUC near 0.5 at SCTE and KSFO), confirming that local seasonal patterns do not transfer across sites. The logistic regression baseline shows moderate zero-shot transferability (AUC 0.79-0.85), suggesting that basic atmospheric state variables contain some universal signal, but falls short of FOG-Net's performance. This gap validates the value of incorporating temporal trends, vertical structure, and radiative forcing information.

A notable observation in Table 5 is that at EGLL, the logistic regression baseline achieves F1=0.47, numerically higher than FOG-Net's F1=0.22, despite FOG-Net's superior performance across all other metrics (AUC: 0.9338 vs 0.8512; MCC: 0.6730 vs 0.4456; AUPRC: 0.5602 vs 0.4023). This apparent contradiction is a threshold calibration artifact: the F1-Score is computed at a fixed threshold (p=0.5), which may favor different models depending on their probability calibration. The threshold-independent AUC metric correctly identifies FOG-Net as having superior discriminative ability—it can better rank fog events above non-fog events across all possible thresholds. This example underscores why threshold-independent metrics (AUC, AUPRC) should be prioritized when comparing models, particularly when models have different probability calibrations or when comparing performance across sites with varying fog climatologies.

# 4. Discussion

## 4.1. The Emergence of Transferable Physics from Data

FOG-Net's zero-shot transfer performance provides evidence for our central hypothesis: the predictability of fog is governed by fundamental physical laws that can be learned from data when a model is constrained to discover transferable mechanisms rather than local patterns. By suppressing geographic coordinates, the model is constrained to identify the underlying physical signal present across all locations.

The SHAP analysis reveals that FOG-Net has reconstructed the known physics of fog formation. The model's convergence on visibility persistence reflects the strong temporal autocorrelation of atmospheric moisture and stability fields—a fundamental property of boundary layer dynamics. Its consistent prioritization of solar angle (angulo_solar) captures the radiative forcing that drives the diurnal cycle of boundary layer evolution, governing both the nocturnal cooling that produces radiation fog and the morning warming that dissipates it. The importance of thermal gradient (gradiente_termico_950_sfc) directly reflects the physical role of temperature inversions in creating the stable stratification necessary for fog persistence, while wind speed (velocidad_viento_10m) captures the competing effects of mixing (beneficial at low speeds) versus turbulent disruption (detrimental at high speeds).

This represents a case of emergent physical discovery, where a data-driven model reconstructs the thermodynamic and radiative mechanisms of a natural phenomenon without being explicitly programmed with the equations of atmospheric physics. The model has learned to evaluate the saturation deficit, assess the strength of radiative cooling, and integrate these factors with the system's recent trajectory—fundamental components of the fog formation process.

## 4.2. The Importance of Temporal Diversity in Training

An earlier Proof-of-Concept (PoC) model trained on a shorter, more recent period (2015–2017, using only 3 years of training data from SCEL) achieved notably lower zero-shot performance compared to the model trained on the extended dataset (2002–2009, using 8 years of training data). The superior performance of the extended-training model across zero-shot validation sites—most notably at KSFO (+0.0673 AUC)—provides evidence that training on a longer, more climatically diverse dataset is important for learning a robust and generalizable representation of atmospheric processes. At the SCEL holdout set, AUC increases modestly (+0.0221) while AUPRC and MCC show mixed changes, reflecting threshold and base-rate effects.

**Table 6: Performance Comparison - PoC Model (3-year training) vs Historical Model (8-year training)**

| Site | Model | Training Period | AUC | AUPRC | MCC | ΔPerformance |
|---|---|---|---|---|---|---|
| SCEL (Holdout) | PoC v1 | 2015-2017 (3 yrs) | 0.9474 | 0.6186 | 0.5612 | Baseline |
| | Historical v2 | 2002-2009 (8 yrs) | 0.9695 | 0.5714 | 0.5324 | +0.0221 AUC |
| SCTE (Zero-Shot) | PoC v1 | 2015-2017 (3 yrs) | 0.9207 | 0.4132 | 0.3152 | Baseline |
| | Historical v2 | 2002-2009 (8 yrs) | 0.9230 | 0.4202 | 0.3324 | +0.0023 AUC |
| KSFO (Zero-Shot) | PoC v1 | 2015-2017 (3 yrs) | 0.8798 | — | — | Baseline |
| | Historical v2 | 2002-2009 (8 yrs) | 0.9471 | 0.0030 | 0.0011 | +0.0673 AUC |

The improvement is most substantial at KSFO (+0.0673 AUC, or 7.6% relative improvement), where the marine advection fog regime differs most from the training site's radiative fog climatology. This suggests that the model trained on the extended 2002–2009 period has learned a more robust representation of the fundamental physical processes. By encompassing a wider range of interannual variability—including diverse El Niño/La Niña phases, varying atmospheric circulation patterns, and different seasonal distributions of fog events—the model can better extrapolate to new meteorological contexts. The extended training period allows the model to capture rare but physically important atmospheric states that may be critical for generalization but are underrepresented in shorter datasets.

### 4.3. A Generalizable Framework for Weather ML

FOG-Net and UTRI-Net (Cerda, 2025) together provide cross-phenomenon validation for a methodological paradigm: coordinate-free, physics-informed ML. While one addresses mesoscale thermodynamics (fog) and the other synoptic-scale dynamics (cyclones), both demonstrate that emphasizing fundamental physical laws and temporal dynamics over static geography is a useful strategy for building transferable weather forecasting models.

### 4.4. Operational Implications and Role as a Decision-Support Tool

FOG-Net is designed as a decision-support tool to complement, not replace, the expertise of human forecasters. Its key operational advantages are consistency, transparency, and the ability to continuously monitor evolving conditions. The model's primary operational focus is on short-term tactical forecasting with lead times of 2–3 hours, which aligns with the typical decision-making window for airport operations including flight scheduling, ground resource allocation, and passenger management. The model's high AUC indicates it can discriminate high-risk from low-risk situations across diverse geographic and climatic settings within this operational timeframe.

**Local Threshold Calibration:** While the model's discriminative ability (AUC) transfers robustly across sites, the raw probability outputs may require site-specific threshold calibration to balance precision and recall according to local operational requirements and fog climatology. For sites with extreme class imbalance such as KSFO (0.03% fog rate), the default 0.5 threshold is suboptimal and produces artificially suppressed classification metrics. We recommend a simple operational calibration protocol: (1) apply the trained model to 1-3 months of local historical data without any retraining, (2) construct a precision-recall curve to visualize the trade-off at different thresholds, and (3) select an optimal threshold based on local cost-benefit analysis of false alarms versus missed events. At EGLL, for example, the default threshold of 0.5 achieved precision of 1.00, suggesting that the model's raw probabilities are well-calibrated for this fog climatology. At sites with lower fog prevalence, lowering the threshold (e.g., to 0.3 or 0.2) will increase recall at the cost of some precision, which may be operationally desirable if the cost of missed fog events is high. This simple post-processing step preserves the model's universal physical reasoning while adapting decision boundaries to local base rates and operational priorities.

The model's explicability, demonstrated through SHAP analysis, fosters trust by allowing forecasters to understand why an alert is being issued and which atmospheric conditions are driving the prediction. This transparency enables human-machine synergy, where forecasters can use the model's assessment as one input among many, while applying their domain expertise to contextualize predictions within the broader meteorological situation.

### 4.5. Limitations and Future Work

**Study limitations include:**

**Geographic scope:** The validation remains limited to four airports, though they span diverse climates and fog regimes. Expansion to additional sites across tropical, polar, and complex terrain environments would further test the boundaries of transferability.

**Data limitations:** METAR observations are subject to spatial discontinuity, as they represent point measurements while fog can be highly localized within airport areas. Additionally, temporal gaps in automated reporting systems can affect data continuity. ERA5 reanalysis, though globally consistent, has inherent limitations: its 0.25° spatial resolution (~28 km at mid-latitudes) may not fully resolve microscale boundary layer processes critical to fog formation, and assimilation system constraints can lead to smoothing of sharp gradients or underestimation of local extremes. The 2002–2009 training period may not fully capture all relevant modes of climate variability, though the comparison with the shorter PoC period (2015–2017) suggests this extended timeframe provides substantial diversity.

**Model architecture:** While XGBoost was chosen for its interpretability and robustness, other architectures (deep learning, probabilistic models) may offer complementary advantages for uncertainty quantification or capturing non-linear interactions at finer scales.

**Baseline comparisons:** This study focuses on demonstrating zero-shot transferability of physics-informed features; direct quantitative comparison with operational NWP guidance or site-specific statistical models was not conducted and remains an important direction for future validation.

**Future work directions:**

1) Expanding validation to more sites across diverse climatic zones, fog regimes, and topographic settings to further test the boundaries of transferability.

2) Conducting systematic quantitative comparison with operational NWP guidance, site-specific statistical models, and persistence baselines to benchmark performance.

3) Evaluating performance sensitivity to data quality issues and missing observations.

4) Exploring ensemble methods or mixture of experts approaches to quantify prediction uncertainty and potentially capture regime-specific behavior.

5) Developing a framework for real-time operational implementation with automated threshold calibration protocols and continuous model monitoring.

6) Investigating the integration of high-resolution topographic information or local circulation features to enhance performance at complex terrain sites.

## 5. Conclusions

This study provides evidence that the fundamental thermodynamic and radiative physics governing fog formation can be transferred across diverse climates and hemispheres when learned through properly constrained machine learning. Through validation across three continents, we have shown that a single coordinate-free model trained on radiative fog in a dry-summer subtropical valley (-33°S) using data from 2002–2009 can successfully discriminate fog-conducive conditions in temperate oceanic climates (-41°S, +51°N) and marine advective regimes (+37°N), maintaining strong skill (average zero-shot AUC of 0.9346) despite distances up to 11,650 km and different meteorological environments.

The model's success is rooted in its reconstruction of physically coherent principles from data. SHAP analysis indicates that FOG-Net has learned to prioritize the atmospheric processes known to govern fog: the persistence of moisture and stability fields, the radiative forcing that drives boundary layer evolution, the thermal inversion strength that maintains stable stratification, and the multi-scale cooling trends that move the atmosphere toward saturation. This convergence on transferable physical predictors—rather than location-specific patterns—supports our hypothesis that fog predictability emerges from fundamental atmospheric physics, not local climatology.

By showing that fundamental atmospheric processes can be discovered and transferred through coordinate-free, physics-informed machine learning, FOG-Net provides a methodological framework and empirical validation for developing geographically applicable atmospheric prediction tools. The model serves as an explicable decision-support system that can complement operational forecasting while advancing our understanding of how data-driven approaches can reconstruct and leverage fundamental physical principles. This work contributes to the goal of developing cost-effective, scientifically grounded tools to enhance aviation safety and demonstrates that machine learning, when properly constrained, can serve as an instrument for discovering and validating transferable atmospheric physics across diverse geographic and climatic regimes.

## Acknowledgments

We thank ECMWF for the ERA5 reanalysis data and Iowa State University for the publicly accessible METAR archive. This independent research received no external funding.

## AI Assistance Disclosure

AI tools were used for data analysis assistance and manuscript drafting support; all conceptual design, methodological decisions, validation strategies, and scientific interpretations remain the sole responsibility of the author.

## Data Availability

The ERA5 reanalysis dataset used in this study is publicly available from the European Centre for Medium-Range Weather Forecasts (ECMWF) Copernicus Climate Data Store at https://cds.climate.copernicus.eu. Historical METAR observations are publicly accessible through the Iowa State University ASOS Network archive at https://mesonet.agron.iastate.edu. The complete source code, including data processing pipeline, model training scripts, evaluation tools, detailed SHAP visualizations (force plots, dependence plots, summary plots), and a comprehensive reproducibility guide with environment specifications, will be made publicly available in the project's GitHub repository within six months of this manuscript's publication or by June 2026, whichever comes first. Researchers interested in early access to the code or collaboration may contact the author directly.

# Appendix A: Analysis of Performance by Forecast Horizon

To assess the model's robustness over time, its performance was evaluated at multiple forecast horizons. This analysis was conducted using the Proof-of-Concept (PoC) model with the 2015–2019 dataset at the SCEL site, prior to development of the final model trained on 2002–2009. FOG-Net is designed as an operational decision-support tool for airport operations, where lead times of 2–3 hours are typically sufficient for executive decision-making regarding flight scheduling, ground operations, and resource allocation. The T+2h and T+3h horizons represent the operationally critical window for airport management. The T+6h horizon was included to demonstrate the graceful degradation pattern and establish the theoretical limits of the physics-informed approach, but extends beyond the primary operational use case. Forecast horizons longer than 6 hours (e.g., T+9h, T+12h, T+24h) fall outside the scope of tactical airport decision support and were not evaluated, as they would require integration with medium-range NWP guidance rather than the short-term, observation-driven framework employed here.

The results show a graceful degradation of performance as the forecast lead time increases, which is consistent with the increasing uncertainty inherent in atmospheric prediction. The analysis of feature importance reveals an adaptation in the model's strategy.

**Table 7: Performance Degradation and Strategic Adaptation by Forecast Horizon (SCEL)**

| Horizon | AUC | AUPRC | MCC | F1-Score (Fog) | Primary Predictive Strategy (from SHAP) |
|---|---|---|---|---|---|
| T+2 hours | 0.9474 | 0.6186 | 0.5612 | 0.59 | Persistence-driven: Relies heavily on visibilidad_actual. |
| T+3 hours | 0.9238 | 0.4300 | 0.4121 | 0.45 | Hybrid: Balances persistence with physical drivers like angulo_solar. |
| T+6 hours | 0.8606 | 0.2604 | 0.2643 | 0.31 | Physics-driven: Relies predominantly on fundamental drivers like angulo_solar and humedad_relativa, as the influence of the initial state fades. |

This strategic shift from relying on persistence for short-term forecasts to relying on fundamental physics for longer-term forecasts demonstrates that FOG-Net is not a simple pattern-matching system, but a model that has learned a physically coherent representation of fog formation dynamics.

# Appendix B: Model Hyperparameters and Training Details

The FOG-Net model is an implementation of the XGBoost (eXtreme Gradient Boosting) algorithm, using the Python xgboost library version 2.0.0. The hyperparameters for the final model (trained on the 2002–2009 period at SCEL) were selected to promote generalization and prevent overfitting, while effectively handling the class imbalance inherent in fog prediction. The key hyperparameters are detailed in Table 8.

**Table 8: Hyperparameters for the Final FOG-Net XGBoost Model**

| Parameter | Value | Justification / Note |
| --- | --- | --- |
| n_estimators | 1000 | The maximum number of boosting rounds. The model was trained for the full duration without early stopping to maximize learning on the extensive historical dataset. |
| learning_rate | 0.05 | A conservative learning rate to ensure stable convergence and reduce the risk of overfitting. |
| max_depth | 5 | Limits the maximum depth of individual trees, preventing the model from learning overly complex and specific patterns. |
| subsample | 0.8 | A form of regularization where each tree is trained on a random 80% subsample of the training data, improving robustness. |
| colsample_bytree | 0.8 | A form of feature regularization where each tree is built using a random 80% subsample of the 19 features. |
| scale_pos_weight | 26.62 | The primary mechanism for handling class imbalance. This value is calculated as the ratio of the number of negative class samples (No Fog) to positive class samples (Fog) in the training set. |
| objective | binary:logistic | Standard objective function for binary classification tasks, outputting probabilities. |
| eval_metric | auc | The Area Under the ROC Curve was used as the evaluation metric to monitor performance during the training process. |

**Training and Preprocessing Protocol**

The training process followed a strict temporal validation methodology. The model was trained on data from 2002 to 2009. A StandardScaler from the scikit-learn library was fitted only on this training partition. This fitted scaler was then saved and used to transform the features of the holdout test set (2010-2012) as well as all zero-shot validation sites (SCTE, KSFO, EGLL). This ensures that no information from the test or validation data leaks into the training process, providing an unbiased estimate of the model's generalization performance.